\def\year{2017}\relax
\documentclass[letterpaper]{article}
\usepackage{aaai17}
\usepackage{graphicx}
\usepackage{balance}  
\usepackage[linesnumbered, ruled, vlined]{algorithm2e}
\usepackage{multirow}
\usepackage[caption=false]{subfig}
\usepackage{overpic}
\usepackage{epstopdf}
\usepackage{color}
\usepackage{tabularx}

\usepackage{adjustbox}

\usepackage{amsmath,amsthm,amssymb,amsfonts}
\usepackage{mathtools}

\DeclareMathOperator*{\argmin}{argmin}
\DeclareMathOperator*{\argmax}{argmax}
\newcommand*{\argminl}{\argmin\limits}
\newcommand*{\argmaxl}{\argmax\limits}
\newcommand{\eat}[1]{}
\mathchardef\figspace=0

\newcommand{\Bx}{\boldsymbol{x}}
\newcommand{\X}{\mathcal{X}}
\newcommand{\Ss}{\mathcal{S}}
\newcommand{\R}{\mathcal{R}}
\newcommand{\T}{\mathcal{T}}
\newcommand{\M}{\mathcal{M}}
\newcommand{\By}{\boldsymbol{y}}
\newcommand{\Bf}{\boldsymbol{f}}
\newcommand{\Ba}{\boldsymbol{\alpha}}
\newcommand{\BQ}{\boldsymbol{Q}}

\newcommand{\cvpic}
{
\put(60, 85){\rotatebox{-20}{1}}
\put(81, 68){\rotatebox{-70}{2}}
\put(88, 41){.}
\put(86, 35){.}
\put(83, 29){.}
\put(59, 16){\rotatebox{-150}{$h$}}
\put(37, 13){\rotatebox{-195}{$h$}}
\put(28, 15){\rotatebox{-205}{+}}
\put(21, 19){\rotatebox{-220}{1}}
\put(8.5, 42){.}
\put(10, 35){.}
\put(13, 30){.}
\put(5, 55){\rotatebox{95}{$k$}}
\put(12, 63){\rotatebox{70}{-}}
\put(10, 67){\rotatebox{60}{1}}
\put(28, 82){\rotatebox{40}{$k$}}
}

\begin{document}
\title{Improving Efficiency of SVM $k$-fold Cross-validation by Alpha Seeding}
\author{Zeyi Wen$^1$, Bin Li$^2$, Kotagiri Ramamohanarao$^1$, Jian Chen$^2$\thanks{Jian Chen is the corresponding author.}, Yawen Chen$^2$, Rui Zhang$^1$\\
$^1$wenzeyi@gmail.com \{rui.zhang, kotagiri\}@unimelb.edu.au\\
The University of Melbourne, Australia\\
$^2$\{gitlinux@gmail.com, ellachen@scut.edu.cn, elfairyhyuk@gmail.com\}\\
South China University of Technology, China}

\maketitle

\begin{abstract}
The $k$-fold cross-validation is commonly used to evaluate the effectiveness of SVMs with the selected hyper-parameters.
It is known that the SVM $k$-fold cross-validation is expensive,
since it requires training $k$ SVMs.
However, little work has explored reusing the $h^{\text{th}}$ SVM for training the $(h+1)^{\text{th}}$ SVM
for improving the efficiency of $k$-fold cross-validation.
In this paper, we propose three algorithms that reuse the $h^{\text{th}}$ SVM
for improving the efficiency of training the $(h+1)^{\text{th}}$ SVM.
Our key idea is to efficiently identify the support vectors and
to accurately estimate their associated weights (also called alpha values) of the next SVM by using the previous SVM.
Our experimental results show that our algorithms are
several times faster than the $k$-fold cross-validation which does not make use of the previously trained SVM.
Moreover, our algorithms produce the same results (hence same accuracy) as the $k$-fold cross-validation
which does not make use of the previously trained SVM.
\end{abstract}

\section{Introduction}
\eat{
Support Vector Machines (SVMs) are widely used in many data mining applications,
such as document classification and image processing~\cite{prasad2010handwriting}.
Training an SVM classifier\footnote{For ease of presentation, we discuss binary classification,
although our approaches are applicable to multi-class classification and regression.}
is to find a hyperplane that separates the two classes of training instances with the largest margin,
such that the SVM classifier can predict the labels of unseen instances more accurately.
\footnote{For ease of presentation, we discuss binary classification,
although our approaches are applicable to multi-class classification and regression.}
\footnote{Without confusion,
we omit ``SVM" in SVM $k$-fold cross-validation.}
}
In order to train an effective SVM classifier,
the hyper-parameters (e.g. the penalty $C$) need to be selected carefully.
The $k$-fold cross-validation is a commonly used process
to evaluate the effectiveness of SVMs with the selected hyper-parameters.
It is known that the SVM $k$-fold cross-validation is expensive,
since it requires training $k$ SVMs with different subsets of the whole dataset.
To improve the efficiency of $k$-fold cross-validation,
some recent studies~\cite{wen2014mascot,athanasopoulos2011gpu} exploit
modern hardware (e.g. Graphic Processing Units).
Chu et al.~\cite{chu2015warm} proposed to reuse the $k$ linear SVM classifiers trained
in the $k$-fold cross-validation with parameter $C$ for training the $k$ linear
SVM classifiers with parameter ($C+\Delta$).
However, little work has explored the possibility of reusing the $h^{\text{th}}$ (where $h \in \{1, 2, ..., (k - 1)\}$) SVM
for improving the efficiency of training the $(h+1)^{\text{th}}$ SVM in the $k$-fold cross-validation
with parameter $C$.

In this paper, we propose three algorithms that reuse the $h^{\text{th}}$ SVM
for training the $(h+1)^{\text{th}}$ SVM in $k$-fold cross-validation.
The intuition behind our algorithms is that the hyperplanes of the two SVMs are similar,
since many training instances (e.g. more than 80\% of the training instances when $k$ is 10) are the same in training the two SVMs.
Note that in this paper we are interested in $k > 2$,
since when $k=2$ the two SVMs share no training instance.

\eat{
Figure~\ref{fig:reuse} gives an example of two SVMs trained during the $k$-fold cross-validation.
In the previous SVM, the optimal hyperplane is shown in Figure~\ref{fig:reuse:before}.
In the training dataset of the next SVM, two instances represented by rectangles (cf. Figure~\ref{fig:reuse:before}) are removed,
and two instances represented by triangles (cf. Figure~\ref{fig:reuse:after}) are added.
The optimal hyperplane of the next SVM is shown using a solid line in Figure~\ref{fig:reuse:after},
where the previous hyperplane is shown in dotted line as a comparison.
As we can see from this example, the hyperplane of the next SVM is potentially close to the previous one,
due to the large number of shared training instances.
Hence, using the previous SVM as a starting point for training the next SVM may save significant amount of computation.

\captionsetup[subfloat]{captionskip=5pt}
\begin{figure}
\center
\subfloat[the previous SVM \label{fig:reuse:before}]{
\begin{overpic}[width=1.4in,height=1.3in]
{fig/jfig/plane-before.eps}
\end{overpic}
}
\subfloat[the next SVM \label{fig:reuse:after}]{
  \begin{overpic}[width=1.4in, height=1.3in]
{fig/jfig/plane-after.eps}
\end{overpic}
}
\caption{Two SVMs in the $k$-fold cross-validation}
\label{fig:reuse}
\end{figure}
}

We present our ideas in the context of training SVMs using Sequential Minimal Optimisation (SMO)~\cite{platt1998sequential},
although our ideas are applicable to other solvers~\cite{osuna1997improved,joachims1999making}.
In SMO, the hyperplane of the SVM is represented by a subset of training instances together with their weights, namely alpha values.
The training instances with alpha values larger than $0$ are called support vectors.
Finding the optimal hyperplane is effectively finding the alpha values for all the training instances.
Without reusing the previous SVM, the alpha values of all the training instances are initialised to $0$.
Our key idea is to use the alpha values of the $h^{\text{th}}$ SVM to initialise the alpha values for the $(h+1)^{\text{th}}$ SVM.
Initialising alpha values using the previous SVM is called \textit{alpha seeding}
in the literature of studying leave-one-out cross-validation~\cite{decoste2000alpha}.
At some risk of confusion to the reader, we will use ``alpha seeding'' and ``initialising alpha values'' interchangeably,
depending on which interpretation is more natural.

Reusing the $h^{\text{th}}$ SVM for training the $(h+1)^{\text{th}}$ SVM in $k$-fold cross-validation has two key challenges.
(i) The training dataset for the $h^{\text{th}}$ SVM is different from that for the $(h+1)^{\text{th}}$ SVM,
but the initial alpha values for the $(h+1)^{\text{th}}$ SVM should be close to their optimal values;
improper initialisation of alpha values leads to slower convergence than without reusing the $h^{\text{th}}$ SVM.
(ii) The alpha value initialisation process should be very efficient;
otherwise, the time spent in the initialisation may be larger than that saved in the training.
This is perhaps the reason that existing work either (i) reuses the $h^{\text{th}}$ SVM trained with parameter $C$ 
for training the $h^{\text{th}}$ SVM with parameter ($C + \Delta$) where both SVMs have the identical training dataset~\cite{chu2015warm}
or (ii) only studies alpha seeding in leave-one-out cross-validation~\cite{decoste2000alpha,lee2004efficient}
which is a special case of $k$-fold cross-validation.

Our key contributions in this paper are the proposal of three algorithms (where we progressively refine one algorithm after the other) 
for reusing the alpha values of the $h^{\text{th}}$ SVM for the $(h+1)^{\text{th}}$ SVM.
(i) Our first algorithm aims to initialise the alpha values to their optimal values for the $(h+1)^{\text{th}}$ SVM
by exploiting the optimality condition of the SVM training.
(ii) To efficiently compute the initial alpha values,
our second algorithm only estimates the alpha values for the newly added instances,
based on the assumption that all the shared instances between the $h^{\text{th}}$ and the $(h+1)^{\text{th}}$ SVMs
tend to have the same alpha values.
(iii) To further improve the efficiency of initialising alpha values,
our third algorithm exploits the fact that a training instance in the $h^{\text{th}}$ SVM
can be potentially replaced by a training instance in the $(h+1)^{\text{th}}$ SVM.
Our experimental results show that when $k=10$, our algorithms are
several times faster than the $k$-fold cross-validation in LibSVM;
when $k=100$, our algorithm dramatically outperforms LibSVM (32 times faster in the Madelon dataset).
Moreover, our algorithms produce the same results (hence same accuracy) as LibSVM.

The remainder of this paper is organised as follows.
We describe preliminaries in Section~\ref{paper:pre}.
Then, we elaborate our three algorithms in Section~\ref{paper:alg},
and report our experimental study in Section~\ref{paper:es}.
In Section~\ref{paper:rw} and~\ref{paper:conc}, we review the related literature,
and conclude this paper.

\section{Preliminaries}
\label{paper:pre}

Here, we give some details of SVMs,
and discuss the relationship of two rounds of $k$-fold cross-validation.

\subsection{Support Vector Machines}
\label{paper:pre-svm}

An instance $\boldsymbol{x}_i$ is attached with an integer $y_i \in \{+1, -1\}$ as its label.
A positive (negative) instance is an instance with the label of $+1$ ($-1$).
Given a set $\mathcal{X}$ of $n$ training instances,
the goal of the SVM training is to find a hyperplane that separates the positive and the
negative training instances in $\mathcal{X}$ with the maximum margin and meanwhile,
with the minimum misclassification error on the training instances.

\eat{
The SVM training is equivalent to solving the following optimisation problem:
\begin{small}
\begin{equation*}
\begin{aligned}
& \underset{\boldsymbol{w}, \text{ } \boldsymbol{\xi}, \text{ } b}{\argminl}
& \frac{1}{2}{||\boldsymbol{w}||^2} + C\sum_{i=1}^{n}{\xi_i} \\
& \text{subject to}
&  y_i(\boldsymbol{w}\cdot \boldsymbol{x}_i + b) \geq 1 - \xi_i \\
& & \xi_i \geq 0, \ \forall i \in \{1,...,n\}
\end{aligned}
\end{equation*}
\end{small}where
$\boldsymbol{w}$ is the normal vector of the hyperplane,
$C$ is the penalty parameter,
$\boldsymbol{\xi}$ is the slack variables
to tolerant some training instances falling into the wrong side of the hyperplane,
and $b$ is the bias of the hyperplane.

In many applications, a better hyperplane exists in other data space instead of the original data space.
}

To enable handily mapping training instances to other data spaces by kernel functions,
finding the hyperplane can be expressed in a \textit{dual form}~\cite{bennett2000duality}
as the following \textit{quadratic programming} problem~\cite{nocedal2006numerical}.
\begin{small}
\vspace{-5pt}
\begin{equation}
\begin{aligned}
& \underset{\Ba}{\argmaxl}
& & \sum_{i=1}^{n}{\alpha_i}-\frac{1}{2}{\Ba^T \BQ \Ba} \\
& \text{subject to}
& &  0 \leq \alpha_i \leq C, \forall i \in \{1,...,n\}; \sum_{i=1}^{n}{y_i\alpha_i} = 0
\end{aligned}
\label{eq:svm_dual}
\end{equation}
\end{small}where 
{$\boldsymbol{\alpha} \in \mathbb{R}^n$} is also called a weight vector,
and $\alpha_i$ denotes the \textit{weight} of $\boldsymbol{x}_i$;
$\boldsymbol{Q}$ denotes an $n \times n$ matrix $[Q_{i,j}]$ and {$Q_{i,j} = y_i y_j K(\boldsymbol{x}_i, \boldsymbol{x}_j)$,
and {$K(\boldsymbol{x}_i, \boldsymbol{x}_j)$} is a kernel value computed from a kernel
function (e.g. Gaussian kernel, {$K(\boldsymbol{x}_i, \boldsymbol{x}_j) =
exp\{-\gamma||\boldsymbol{x}_i-\boldsymbol{x}_j||^2\}$}).
Then, the  goal of the SVM training is to find the optimal $\Ba$.
If $\alpha_i$ is greater than $0$, $\Bx_i$ is called a \textit{support vector}.

\eat{An intriguing property of the SVM is that the SVM classifier is determined by the support vectors,
and the non-support vectors have no effect on the SVM classifier.

The key ideas of our algorithm proposed in Section~\ref{paper:alg} are applicable to various SVM training algorithms
such as chucking based approaches~\cite{osuna1997improved,joachims1999making},
Sequential Minimal Optimisation (SMO)~\cite{platt1998sequential},
and training algorithms in the SVM primal form~\cite{shalev2011pegasos,suykens1999least}.
}

In this paper, we present our ideas in the context of
using SMO to solve Problem~\eqref{eq:svm_dual},
although our key ideas are applicable to other solvers~\cite{osuna1997improved,joachims1999making}.
The training process and the derivation of the optimality condition are unimportant for understanding our algorithms,
and hence are not discussed here.
Next, we present the optimality condition for the SVM training which
will be exploited in our proposed algorithms in Section~\ref{paper:alg}.

\subsubsection{The optimality condition for the SVM training}
In SMO, a training instance $\Bx_i$ is associated with an optimality indicator $f_i$ which is defined as follows.
\begin{small}
\vspace{-5pt}
\begin{equation}
\vspace{-5pt}
f_i = y_i\sum_{j=1}^{n}{\alpha_j Q_{i,j} - y_i}
\label{eq:f-i}
\end{equation}
\end{small}The optimality condition of the SVM training is the Karush-Kuhn-Tucker (KKT)~\cite{kuhn2014nonlinear} condition.
When the optimality condition is met,
we have the optimality indicators satisfying the following constraint.
\begin{small}
\begin{equation}
\min\{f_i| i \in I_u \cup I_m\} \ge
\max\{f_i | i \in I_l \cup I_m\}
\label{eq:fu-fl}
\vspace{-5pt}
\end{equation}
\end{small}where
\begin{small}
\begin{equation}
\begin{adjustbox}{max width=0.43\textwidth}
$
\begin{split}
&I_{m} = \{i | \boldsymbol{x}_i \in \mathcal{X}, 0 < \alpha_i < C\},\\
&I_{u} = \{i | \boldsymbol{x}_i \in \mathcal{X}, y_i = +1, \alpha_i = 0\} \cup \{i | \boldsymbol{x}_i \in \mathcal{X}, y_i = -1, \alpha_i = C\},\\
&I_{l} = \{i | \boldsymbol{x}_i \in \mathcal{X}, y_i = +1, \alpha_i = C\} \cup \{i | \boldsymbol{x}_i \in \mathcal{X}, y_i = -1, \alpha_i = 0\}.
\end{split}
$
\end{adjustbox}
\label{eq:def-i}
\end{equation}
\end{small}As
observed by Keerthi et al.~\cite{keerthi2001improvements},
Constraint~\eqref{eq:fu-fl} is equivalent to the following constraints.
\begin{small}
\begin{equation}
f_i > b \text{ for } i \in I_u; \text{\hspace{5pt}} f_i = b \text{ for } i \in I_m; \text{\hspace{5pt}} f_i < b \text{ for } i \in I_l
\text{\hspace{5pt}}
\label{eq:f-const}
\end{equation}
\end{small}where
$b$ is the bias of the hyperplane.
Our algorithms proposed in Section~\ref{paper:alg} exploit Constraint~\eqref{eq:f-const}.

\captionsetup[subfloat]{captionskip=5pt}
\begin{figure}
\vspace{-10pt}
\center
\subfloat[$k$ subsets \label{fig:cv:kfold}]{
\begin{overpic}[width=1in,height=1in]
{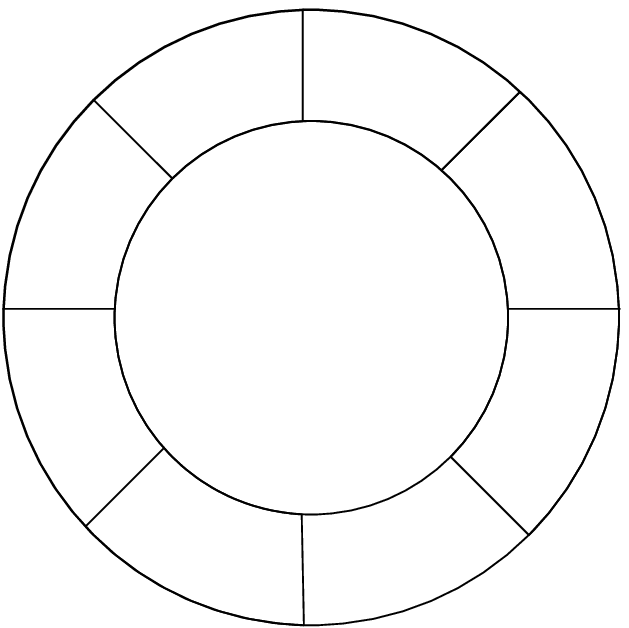}
\cvpic
\end{overpic}
}
\subfloat[$h^{\text{th}}$ as test \label{fig:cv:1stfold}]{
  \begin{overpic}[width=1in, height=1in]
{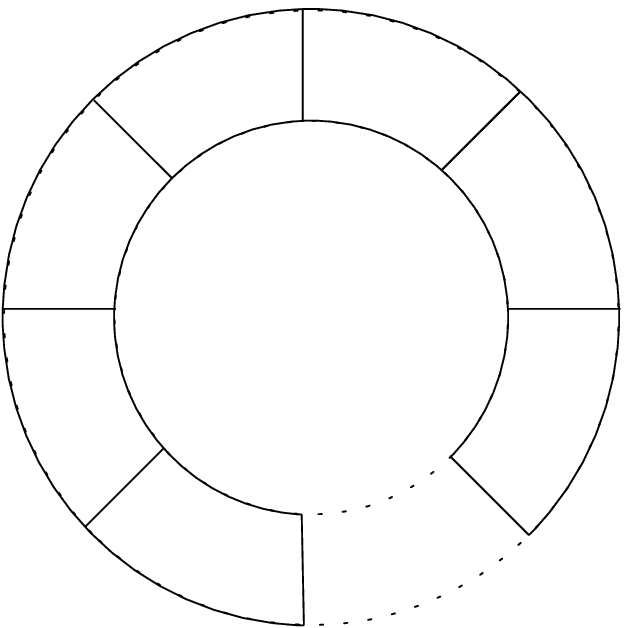}
\cvpic
\end{overpic}
}
\subfloat[$(h+1)^{\text{th}}$ as test\label{fig:cv:pthfold}]{
  \begin{overpic}[width=1in, height=1in]
{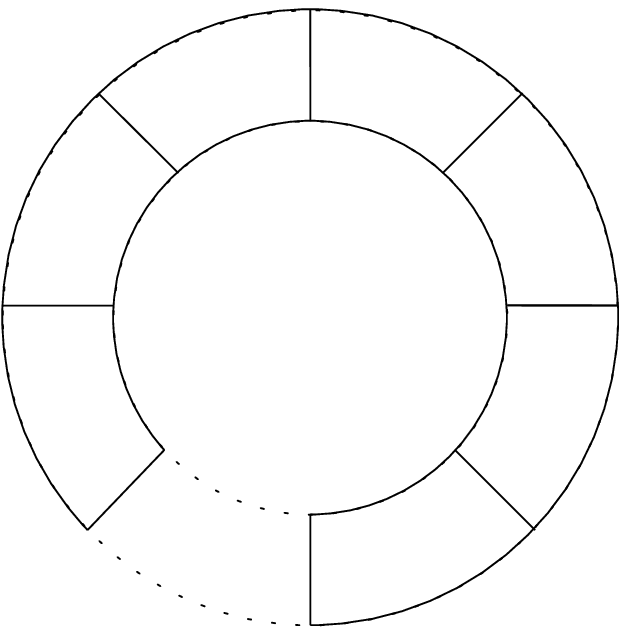}
\cvpic
\end{overpic}
}
\vspace{-5pt}
\caption{$k$-fold cross-validation}
\vspace{-10pt}
\label{fig:cv}
\end{figure}

\eat{
\subsection{$k$-fold cross-validation}
The $k$-fold cross-validation evenly divides the dataset into $k$ subsets.
One subset is used as the test set $\mathcal{T}$,
while the rest $(k-1)$ subsets together form the training set $\mathcal{X}$.
The SVM is first trained using $\mathcal{X}$.
Then the trained SVM is used to classify the test instances in $\mathcal{T}$ by predicting their labels.
To obtain more reliable results, the above training-classification process is repeated for $k$ times,
where every subset is used as the test set in turn.
Figure~\ref{fig:cv} shows an example of the $k$-fold cross-validation.
The whole dataset is evenly divided into $k$ subsets as shown in Figure~\ref{fig:cv:kfold}.
In the first round, the $1^{st}$ subset is used as the test set
and the remaining $(k-1)$ subsets are used as the training set (cf. Figure~\ref{fig:cv:1stfold});
similarly in the $h^{th}$ round, the $h^{th}$ subset is used as the test set
and the remaining $(k-1)$ subsets are used as the training set (cf. Figure~\ref{fig:cv:pthfold}).
As we can see from the $k$-fold cross-validation process,
$(k-2)$ subsets of the training instances are shared in any two rounds.
Note that we are interested in the $k$-fold cross-validation where $k > 2$,
since no instance is shared when $k=2$.
}

\subsection{Relationship between the $h^{\text{th}}$ round and the $(h+1)^{\text{th}}$ round in $k$-fold cross-validation}
\label{paper:relation-two-round}
The $k$-fold cross-validation evenly divides the dataset into $k$ subsets.
One subset is used as the test set $\mathcal{T}$,
while the rest $(k-1)$ subsets together form the training set $\mathcal{X}$.
Suppose we have trained the $h^{\text{th}}$ SVM (in the $h^{\text{th}}$ round)
using the $1^{\text{st}}$ to $(h-1)^{\text{th}}$ and $(h+1)^{\text{th}}$ to $k^{\text{th}}$ subsets as the training set,
and the $h^{\text{th}}$ subset serves as the testing set (cf. Figure~\ref{fig:cv:1stfold}).
Now we want to train the $(h+1)^{\text{th}}$ SVM.
Then, the $1^{\text{st}}$ to $(h-1)^{\text{th}}$ subsets and the $(h+2)^{\text{th}}$ to $k^{\text{th}}$
subsets are shared between the two rounds of the training.
To convert the training set used in the $h^{\text{th}}$ round to the training set for the $(h+1)^{\text{th}}$ round,
we just need to remove the $(h+1)^{\text{th}}$ subset from and add the $h^{\text{th}}$ subset
to the training set used in the $h^{\text{th}}$ round.
Hereafter, we call the $h^{\text{th}}$ and $(h+1)^{\text{th}}$ SVMs \textit{the previous SVM} and \textit{the next SVM}, respectively.

For ease of presentation, we denote the shared subsets---$(k - 2)$ subsets in total---by $\Ss$,
denote the unshared subset in the training of the previous round by $\R$,
and denote the subset for testing in the previous round by $\T$.
Let us continue to use the example shown in Figure~\ref{fig:cv},
$\Ss$ consists of the $1^{\text{st}}$ to $(h-1)^{th}$ subsets and the $(h+2)^{\text{th}}$ to $k^{\text{th}}$ subsets;
$\R$ is the $(h+1)^{\text{th}}$ subset; $\T$ is the $h^{\text{th}}$ subset.
To convert the training set $\X$ used in the $h^{\text{th}}$ round to the training set $\X'$ for the $(h+1)^{\text{th}}$ round,
we just need to remove $\R$ from $\X$ and add $\T$ to $\X$, i.e. $\X' = \T \cup \X \setminus \R = \T \cup \Ss$.
We denote three sets of indices as follows corresponding to
$\R$, $\T$ and $\Ss$ by $I_\R$, $I_\T$ and $I_\Ss$, respectively.
\begin{small}
\begin{equation}
I_{\R} = \{i | \Bx_i \in \R\}, I_{\T} = \{i | \Bx_i \in \T\}, I_{\Ss} = \{i | \Bx_i \in \Ss\}
\label{eq:i-rt}
\vspace{-10pt}
\end{equation}
\end{small}

Two rounds of the $k$-fold cross-validation often have many training instances in common, i.e. large $\Ss$.
E.g. when $k$ is 10, $\frac{8}{9}$ (or $\sim 90\%$) of instances in $\X$ and $\X'$ are the instances of $\Ss$.
Next, we study three algorithms for reusing the previous SVM to train the next SVM.

\section{Reusing the previous SVM in $k$-fold cross-validation}
\label{paper:alg}
We present three algorithms that reuse the previous SVM for training the next SVM,
where we progressively refine one algorithm after the other.
(i) Our first algorithm aims to initialise the alpha values $\Ba'$ to their optimal values for the next SVM,
based on the alpha values $\Ba$ of the previous SVM.
We call the first algorithm Adjusting Alpha Towards Optimum (ATO).
(ii) To efficiently initialise $\Ba'$, our second algorithm keeps the alpha values
of the instances in $\Ss$ unchanged (i.e. $\alpha_s' = \alpha_s$ for $s \in I_\Ss$),
and estimates $\alpha_t'$ for $t \in I_\T$.
This algorithm effectively performs alpha value initialisation via replacing $\R$ by $\T$
under constraints of Problem~\eqref{eq:svm_dual},
and hence we call the algorithm Multiple Instance Replacement (MIR).
(iii) Similar to MIR, our third algorithm also keeps the alpha values of the instances in $\Ss$ unchanged;
different from MIR, the algorithm replaces the instances in $\R$ by the instances in $\T$ one at a time,
which dramatically reduces the time for initialising $\Ba'$.
We call the third algorithm Single Instance Replacement (SIR).
Next, we elaborate these three algorithms.

\subsection{Adjusting Alpha Towards Optimum (ATO)}

ATO aims to initialise the alpha values to their optimal values.
It employs the technique for online SVM training, designed by Karasuyama and Takeuchi~\cite{karasuyama2009multiple},
for the $k$-fold cross-validation.
In the online SVM training, a subset $\R$ of outdated training instances is removed from the training set $\X$,
i.e. $\X' = \X \setminus \R$;
a subset $\T$ of newly arrived training instances  is added to the training set, i.e. $\X' = \X' \cup \T$.
The previous SVM trained using $\X$ is adjusted by removing and adding subsets of instances to obtain the next SVM.

In the ATO algorithm, we first construct a new training dataset $\X'$ where $\X' = \Ss = \X \setminus \R$.
Then, we gradually increase alpha values of the instances in $\T$ (i.e. increase $\alpha_t'$ for $t \in I_\T$), denoted by $\Ba_\T'$,  to (near) their optimal values;
meanwhile, we gradually decrease the alpha values of the instances in $\R$ (i.e. decrease $\alpha_r'$ for $r \in I_\R$),
denoted by $\Ba_\R'$, to $0$.
Once the alpha value of an instance in $\T$ satisfies the optimal condition (i.e. Constraint~\eqref{eq:f-const}),
we move the instance from $\T$ to the training set $\X'$;
similarly once the alpha value of an instance in $\R$ equals to 0 (becoming a non-support vector),
we remove the instance from $\R$.
ATO terminates the alpha value initialisation when $\R$ is empty.

\subsubsection{Updating the alpha values}
Next, we present details of increasing $\Ba_\T'$ and decreasing $\Ba_\R'$.
We denote the step size for an increment on $\Ba_\T'$ and decrement on $\Ba_\R'$ by $\eta$.
From constraints of Problem~\eqref{eq:svm_dual}, all the alpha values must be in $[0, C]$.
Hence, for $t \in I_\T$ the increment of $\alpha_t'$, denoted by $\Delta \alpha_t'$, cannot exceed $(C - \alpha_t')$;
for $r \in I_\R$ the decrement of $\alpha_r'$, denoted by $\Delta \alpha_r'$, cannot exceed $\alpha_r'$.
We denote the change of all the alpha values of the instances in $\T$ by $\Delta \Ba_\T'$ and
the change of all the alpha values of the instances in $\R$ by $\Delta \Ba_\R'$.
Then, we can compute $\Delta \Ba_\T'$ and $\Delta \Ba_\R'$ as follows.
\begin{small}
\begin{equation}
\Delta \Ba_\T'= \eta (C\boldsymbol{1} - \Ba_\T'), \text{\hspace{10pt}} \Delta \Ba_\R' = -\eta \Ba_\R'
\label{eq:delta-a-rt}
\end{equation}
\end{small}where $\boldsymbol{1}$ is a vector with all the dimensions of $1$.
When we add $\Delta \Ba_\T'$ to $\Ba_\T'$ and $\Delta \Ba_\R'$ to $\Ba_\R'$,
constraints of Problem~\eqref{eq:svm_dual} must be satisfied.
However, after adjusting $\Ba_\T'$ and $\Ba_\R'$, the constraint $\sum_{i \in I_\T \cup I_\Ss \cup I_\R}{y_i \alpha_i'} = 0$ is often violated,
so we need to adjust the alpha values of the training instances in $\X'$
(recall that at this stage $\X' = \Ss$).
We propose to adjust the alpha values of the training instances in $\X'$ which are also in $\M$
where $\Bx_i \in \M$ given $i \in I_m$.
\eat{The reason for choosing $\M$ is that the alpha values can be increased or decreased under the constraint $\alpha \in [0, C]$.
}In summary, after increasing $\Ba_\T'$ and decreasing $\Ba_\R'$, we adjust $\Ba_\M'$.
So when adjusting $\Ba_\T'$, $\Ba_\R'$ and $\Ba_\M'$, we have the following equation according to constraints of Problem~\eqref{eq:svm_dual}.
\begin{small}
\begin{equation}
\sum_{t \in I_\T}y_t \Delta \alpha_t' +\sum_{r \in I_\R}y_r \Delta \alpha_r' +\sum_{i \in I_m}y_i \Delta \alpha_i' =0
\label{eq:after-adj-alpha-sum}
\end{equation}
\end{small}$\M$ often has a large number of instances, and there are many possible ways to adjust $\Ba_\M'$.
Here, we propose to use the adjustment on $\Ba_\M'$ that ensures
all the training instances in $\M$ satisfy the optimality condition (i.e. Constraint~\eqref{eq:f-const}).
According to Constraint~\eqref{eq:f-const}, we have $\forall i \in I_m$ and $f_i = b$.
Combining $f_i = b$ and the definition of $f_i$ (cf. Equation~\eqref{eq:f-i}), we have the following equation for each $i \in I_m$.
\begin{small}
\begin{equation}
y_i(\sum_{t \in I_\T} Q_{i,t}\Delta\alpha_t' +\sum_{r \in I_\R} Q_{i,r}\Delta\alpha_r' +\sum_{j \in I_m} Q_{i,j}\Delta\alpha_j')=0
\label{eq:after-adj}
\end{equation}
\end{small}Note that $y_i$ can be omitted in the above equation.
We can rewrite Equation~\eqref{eq:after-adj-alpha-sum} and Equation~\eqref{eq:after-adj} using
the matrix notation for all the training instances in $\M$.
\begin{small}
\begin{equation*}
\begin{bmatrix}
\By_\T^T		& \By_\R^T\\
\BQ_{\M,\T}	& \BQ_{\M,\R}
\end{bmatrix}
\begin{bmatrix}
\Delta \Ba_\T'\\
\Delta \Ba_\R'
\end{bmatrix}+
\begin{bmatrix}
\By_\M^T\\
\BQ_{\M,\M}
\end{bmatrix}
\Delta\Ba_M'
=0
\end{equation*}
\end{small}We substitute $\Delta \Ba_\T'$ and $\Delta \Ba_\R'$ using Equation~\eqref{eq:delta-a-rt};
the above equation can be rewritten as follows.
\begin{small}
\begin{equation}
\boldsymbol{\Delta \alpha}_\M' = -\eta \Phi
\label{eq:delta-a-m}
\end{equation}
\end{small}where {\small $\Phi = \begin{bmatrix}
	\By_\M^T\\
	\BQ_{\M, \M}
\end{bmatrix}^{-1}
\begin{bmatrix}
	\By^T_\T   & \By^T_\R \\
	\BQ_{\M,\T} & \BQ_{\M,\R}
\end{bmatrix}
\begin{bmatrix}
	C\boldsymbol{1} - \Ba_\T' \\ 
	-\Ba_\R'
\end{bmatrix}$}.
If the inverse of the matrix in Equation~\eqref{eq:delta-a-m} does not exist,
we find the pseudo inverse~\cite{greville1960some}

\eat{
Note that the set $\M$ may be empty in some cases. 
When $\M$ is empty, we randomly pick one instance from $I_u$ and one from $I_l$.
We then set the alpha values of the four instances to $C/2$, and put them to $\M$.
Recall that the ideal alpha value $\alpha_m'$ of the instance in $\M$ is $0 < \alpha_m' < C$.
}

\textbf{Computing step size $\eta$}:
Given an $\eta$,
we can use Equations~\eqref{eq:delta-a-rt} and~\eqref{eq:delta-a-m} to adjust $\Ba_\M'$, $\Ba_\T'$ and $\Ba_\R'$.
The changes of the alpha values lead to the change of all the optimality indicators $\Bf$.
We denote the change to $\Bf$ by $\Delta \Bf$ which can be computed by the following equation derived from Equation~\eqref{eq:f-i}.
\begin{small}
\begin{equation}
\boldsymbol{y} \odot \boldsymbol{\Delta f} = \eta [-\BQ_{\X, \M} \Phi+ \BQ_{\X,\T}(C\boldsymbol{1} - \Ba_\T') - \BQ_{\X,\R}\Ba_\R']
\label{eq:delta-f}
\end{equation}
\end{small}where $\odot$ is the hadamard product (i.e. element-wise product~\cite{schott2005matrix}).\

If the step size $\eta$ is too large, more optimality indicators tend to violate Constraint~\eqref{eq:f-const}.
Here, we use Equation~\eqref{eq:delta-f} to compute the step size $\eta$
by letting the updated $f_i$ (where $i \in I_u \cup I_l$) just violate Constraint~\eqref{eq:f-const},
i.e. $f_i + \Delta f_i = b$ for $i \in I_u \cup I_l$.

\eat{
The $\eta$ computed from Equation~\eqref{eq:delta-f} may need to be reduced,
because the $\eta$ will be used to compute $\Delta \Ba_\T'$, $\Delta \Ba_\R'$ and $\Delta \Ba_\M'$ (cf. Equations~\eqref{eq:delta-a-rt} and~\eqref{eq:delta-a-m})
and the updated $\Ba'$ should satisfy $\alpha_i' \in [0, C]$ for all $i \in I_\T \cup I_\R \cup I_m$.
After a valid $\eta$ is obtained, we update $\Ba'$ using Equations~\eqref{eq:delta-a-rt} and~\eqref{eq:delta-a-m}.
}

\subsubsection{Updating $\Bf$}
After updating $\Ba'$,
we update $\Bf$ using Equations~\eqref{eq:f-i} and~\eqref{eq:delta-f}.
Then, we update the sets $I_m$, $I_u$ and $I_l$ according to Constraint~\eqref{eq:f-const}.

The process of computing $\eta$ and updating $\Ba'$ and $\Bf$ are repeated until $\R$ is empty.
\subsubsection{Termination}
When $\R$ is empty, the SVM may not be optimal,
because the set $\T$ may not be empty.
The alpha values obtained from the above process serve as the initial alpha values for the next SVM.
To obtain the optimal SVM, we use SMO to adjust the initial alpha values until optimal condition is met.
The pseudo-code of the full algorithm is shown in Algorithm~\ref{alg:ato} in Supplementary Material.
\eat{
until the optimality condition (cf. Constraints~\eqref{eq:def-i} and~\eqref{eq:f-const}) is satisfied.

According to the definition (cf. Equation~\eqref{eq:def-i}), $I_m$, $I_u$ and $I_l$ are determined
by alpha values and their labels of the training instances.
Note that we cannot make $I_m$, $I_u$ and $I_l$ satisfy both Equation~\eqref{eq:def-i} and Constraint~\eqref{eq:f-const}
(i.e. the optimality condition),
since after updating $\Bf$ and $\Ba'$ the KKT condition is often violated.
To continue to reduce $\Ba_\R'$ until $\R$ becomes empty,
we rearrange $I_m$, $I_u$ and $I_l$ by only considering the sign of $(f_i - b)$.}

\subsection{Multiple Instance Replacement (MIR)}

A limitation of ATO is that
it requires adjusting \textit{all} the alpha values for an \textit{unbounded} number of times (i.e. until $\R$ is empty).
Hence, the cost of initialising the alpha values may be very high.
In what follows, we propose the Multiple Instance Replacement (MIR) algorithm that only needs to adjust $\Ba_\T'$ once.
The alpha values of the shared instances between the two rounds stay unchanged
(i.e. $\Ba_\Ss' = \Ba_\Ss$), the intuition is that many support vectors tend to stay unchanged.
The key idea of MIR is to replace $\R$ by $\T$ at once.

We obtain the alpha values of the instances in $\Ss$ and $\R$ from the previous SVM,
and those alpha values satisfy the following constraint.
\begin{small}
\vspace{-5pt}
\begin{equation}
\sum_{s \in I_\Ss}{y_s\alpha_s} + \sum_{r \in I_\R}{y_r\alpha_r} = 0
\label{eq:alpha_sum_const}
\vspace{-5pt}
\end{equation}
\end{small}

In the next round of SVM $k$-fold cross-validation, $\R$ is removed and $\T$ is added.
When reusing alpha values, we should guarantee that the above constraint holds.
To improve the efficiency of initialising alpha values, we do not change alpha values in first term
of Constraint~\eqref{eq:alpha_sum_const}, i.e. $\sum_{s \in I_\Ss}{y_s\alpha_s}$.

To satisfy the above constraint after replacing $\R$ by $\T$,
we only need to ensure $\sum_{r \in I_\R}{y_r\alpha_r} = \sum_{t \in I_\T}{y_t\alpha_t'}$.
Next, we present an approach to compute $\Ba_\T'$.

According to Equation~\eqref{eq:f-i}, we can rewrite $f_i$ before replacing $\R$ by $\T$ as follows.
\begin{small}
\vspace{-5pt}
\begin{equation}
f_i =
y_i(\sum_{r \in I_\R}{\alpha_r Q_{i, r}} +
\sum_{s \in I_\Ss}{\alpha_s Q_{i, s}} - 1)
\label{eq:f-i-before}
\vspace{-5pt}
\end{equation}
\end{small}After replacing $\R$ by $\T$, $f_i$ can be computed as follows.
\begin{small}
\vspace{-5pt}
\begin{equation}
f_i = y_i(\sum_{t \in I_\T}{\alpha_t' Q_{i, t}} + \sum_{s \in I_\Ss}{\alpha_s' Q_{i, s}} - 1)
\label{eq:f-i-after}
\vspace{-5pt}
\end{equation}
\end{small}where $\alpha_s' = \alpha_s$, i.e. the alpha values in $\Ss$ stay unchanged.
We can compute the change of $f_i$, denoted by $\Delta f_i$, by subtracting Equation~\eqref{eq:f-i-before} from Equation~\eqref{eq:f-i-after}.
Then, we have the following equation.
\begin{small}
\vspace{-5pt}
\begin{equation}
\Delta f_i = y_i[\sum_{t \in I_\T}{\alpha_t' Q_{i, t}} -
\sum_{r \in I_\R}{\alpha_r Q_{i, r}}]
\label{eq:delta-f-const}
\vspace{-5pt}
\end{equation}
\end{small}To meet the constraint
$\sum_{}{y_i \alpha_i = 0}$ after replacing $\R$ by $\T$, we have the following equation.
\begin{small}
\vspace{-5pt}
\begin{equation*}
\sum_{s \in I_\Ss}{y_s\alpha_s} + \sum_{r \in I_\R}{y_r\alpha_r} = \sum_{s \in I_\Ss}{y_s\alpha_s'} + \sum_{t \in I_\T}{y_t \alpha_t'}
\vspace{-5pt}
\end{equation*}
\end{small}As $\alpha_s' = \alpha_s$, we rewrite the above equation as follows.
\begin{small}
\vspace{-5pt}
\begin{equation}
\sum_{r \in I_\R}{y_r\alpha_r} = \sum_{t \in I_\T}{y_t \alpha_t'}
\label{eq:delta-alpha-const}
\vspace{-5pt}
\end{equation}
\end{small}We write Equations~\eqref{eq:delta-f-const} and~\eqref{eq:delta-alpha-const} together
as follows.
\begin{small}
\vspace{-5pt}
\begin{equation}
\begin{bmatrix}
\boldsymbol{y} \odot \boldsymbol{\Delta f} + \boldsymbol{Q}_{\X,\R}\boldsymbol{\alpha}_{\R}\\
\boldsymbol{y}_{\R}^T \cdot \boldsymbol{\alpha}_{\R}
\end{bmatrix}
= 
\begin{bmatrix}
\boldsymbol{Q}_{\X,\T} \\ \boldsymbol{y}_{\T}^T
\end{bmatrix}
\Ba'_{\T}
\label{eq:delta-f2}
\end{equation}
\end{small}Similar to the way we compute $\Delta f_i$ in the ATO algorithm,
given $i$ in $I_u \cup I_l$ we compute $\Delta f_i$ by letting $f_i + \Delta f_i = b$ (cf. Constraint~\eqref{eq:f-const}).
Given $i$ in $I_m$, we set $\Delta f_i = 0$ since we try to avoid $f_i$ violating Constraint~\eqref{eq:f-const}.
Once we have $\Delta \Bf$, the only unknown in Equation~\eqref{eq:delta-f2} is $\Ba_{\T}'$.

\eat{
Note that for $i \in I_\R$, $\Delta f_i$ need not to be included in Equation~\eqref{eq:delta-f2},
since $\Bx_i$ will be removed anyway and hence $\Delta f_i$ can be any value.
}

\subsubsection{Finding an approximate solution for $\Ba_\T'$}

The linear system shown in Equation~\eqref{eq:delta-f2} may have no solution.
This is because $\Ba_\Ss'$ may also need to be adjusted,
but is not considered in Equation~\eqref{eq:delta-f2}.
Here, we propose to find the approximate solution $\Ba_{\T}'$ for Equation~\eqref{eq:delta-f2}
by using linear least squares~\cite{lawson1974solving} and we have the following equation.
\begin{small}
\vspace{-5pt}
\begin{equation*}
\begin{bmatrix}
	\BQ_{\X,\T} \\ \By_{\T}^T
\end{bmatrix}^T 
\begin{bmatrix}
	\By \odot \boldsymbol{\Delta f} + \BQ_{\X,\R}\Ba_{\R}\\
	\By_{\R}^T \cdot \Ba_{\R}
\end{bmatrix}
=
\begin{bmatrix}
\BQ_{\X, \T} \\ \By_{\T}^T
\end{bmatrix}^T 
\begin{bmatrix}
\BQ_{\X, \T} \\ \By_{\T}^T
\end{bmatrix}
\Ba_{\T}'
\end{equation*}
\vspace{-5pt}
\end{small}Then we can compute $\Ba_\T'$ using the following equation.
\begin{equation}
\begin{adjustbox}{max width=0.41\textwidth}
$\Ba_{\T}'=
\Big(
\begin{bmatrix}
\BQ_{\X, \T} \\ \By_{\T}^T
\end{bmatrix}^T 
\begin{bmatrix}
\BQ_{\X, \T} \\ \By_{\T}^T
\end{bmatrix}
\Big)^{-1}
\begin{bmatrix}
	\BQ_{\X,\T} \\ \By_{\T}^T
\end{bmatrix}^T 
\begin{bmatrix}
	\By \odot \boldsymbol{\Delta f} + \BQ_{\X,\R}\Ba_{\R}\\
	\By_{\R}^T \cdot \Ba_{\R}
\end{bmatrix}$
\label{eq:delta-alpha-appr}
\end{adjustbox}
\end{equation}
If the inverse of the matrix in above equation does not exist,
we find the pseudo inverse similar to ATO.

\subsubsection{Adjusting $\Ba_\T'$}
\label{paper:cv-rpi-adjust}
Due to the approximation,
the constraints $0 \le \alpha_t' \le C$ and
$\sum_{r \in I_\R}{y_r\alpha_r} = \sum_{t \in I_\T}{y_t \alpha_t'}$ may not hold.
Therefore, we need to adjust $\Ba_\T'$ to satisfy the constraints, and we perform the following steps.
\begin{itemize}
	\item If $\alpha_t' < 0$, we set $\alpha_t' = 0$;
		  if $\alpha_t' > C$, we set $\alpha_t' = C$.
	\item If {\footnotesize $\sum_{t \in I_\T}{y_t \alpha_t'} > \sum_{r \in I_\R}{y_r\alpha_r}$ }
		  (if {\footnotesize $\sum_{t \in I_\T}{y_t \alpha_t'} < \sum_{r \in I_\R}{y_r\alpha_r}$}),
		  we uniformly decrease (increase) all the $y_t \alpha_t'$ until $\sum_{t \in I_\T}{y_t \alpha_t'} = \sum_{r \in I_\R}{y_r\alpha_r}$,
		  subjected to the constraint $0 \le \alpha_t' \le C$.
\end{itemize}

After the above adjusting, $\alpha_t'$ satisfies the constraints $0 \le \alpha_t' \le C$
and $\sum_{r \in I_\R}{y_r\alpha_r} = \sum_{t \in I_\T}{y_t \alpha_t'}$.
Then, we use SMO with $\Ba'$ (where $\Ba' = \Ba_\Ss' \cup \Ba_\T'$) as the initial alpha values
for training an optimal SVM.
The pseudo-code of whole algorithm is shown in Algorithm~\ref{alg:rt-replacement} in Supplementary Material.

\subsection{Single Instance Replacement (SIR)}

Both ATO and MIR have the following major limitation:
the computation for $\Ba_\T'$ is expensive (e.g. require computing the inverse of a matrix).
The goal of the ATO and MIR is to minimise the number of instances that violate the optimality condition.
In the algorithm we propose here, we try to \textit{minimise} $\Delta f_i$ with a hope that the small change to $f_i$
will not violate the optimality condition.
This slight change of the goal leads to a much cheaper computation cost on computing $\Ba_\T'$.
Our key idea is to replace the instance in $\R$ one after another with a similar instance in $\T$.
Since we replace one instance in $\R$ by an instance in $\T$ each time,
we call this algorithm Single Instance Replacement (SIR).
Next, we present the details of the SIR algorithm.

According to Equation~\eqref{eq:f-i}, we can rewrite $f_i$ of the previous SVM as follows.
\begin{small}
\vspace{-5pt}
\begin{equation}
f_i = y_i(\sum_{j \in I_\Ss \cup I_\R \setminus \{p\}}{\alpha_j Q_{i, j}} +
	  \alpha_p Q_{i, p} - 1)
\label{eq:before-single-replace}
\vspace{-5pt}
\end{equation}
\end{small}where $p \in I_\R$. We replace the training instance $\Bx_p$ by $\Bx_q$ where $q \in I_\T$,
and then the value of $f_i$ after replacing $\Bx_p$ by $\Bx_q$ is as follows.
\begin{small}
\vspace{-5pt}
\begin{equation}
f_i = y_i(\sum_{j \in I_\Ss \cup I_\R \setminus \{p\}}{\alpha_j Q_{i, j}} +
	  \alpha_q' Q_{i, q} - 1)
\label{eq:after-single-replace}
\vspace{-5pt}
\end{equation}
\end{small}where $\alpha_q' = \alpha_p$.
By subtracting Equation~\eqref{eq:before-single-replace} from Equation~\eqref{eq:after-single-replace},
the change of $f_i$, denoted by $\Delta f_i$, can be computed by 
$\Delta f_i = y_i \alpha_p (Q_{i, q} - Q_{i, p})$.
Recall that $Q_{i, j} = y_i y_j K(\Bx_i, \Bx_j)$. We can write $\Delta f_i$ as follows.
\vspace{-3pt}
\begin{equation}
\vspace{-3pt}
\Delta f_i = \alpha_p (y_q K(\Bx_i, \Bx_q) - y_p K(\Bx_i, \Bx_p))
\end{equation}
Recall also that in SIR we want to replace $\Bx_p$ by an instance, denoted by $\Bx_q$, that minimises $\Delta f_i$.
When $\alpha_p = 0$, $\Delta f_i$ has no change after replacing $\Bx_p$ by $\Bx_q$.
In what follows, we focus on the case that $\alpha_p > 0$.

We propose to replace $\Bx_p$ by $\Bx_q$ if $\Bx_q$ is the ``most similar'' instance to $\Bx_p$ among all the instances in $\T$.
The instance $\Bx_q$ is called the most similar to the instance $\Bx_p$ among all the instances in $\T$,
when the following two conditions are satisfied.
\begin{itemize}
	\item $\Bx_p$ and $\Bx_q$ have the same label, i.e. $y_p = y_q$.
	\item $K(\Bx_p, \Bx_q) \ge K(\Bx_p, \Bx_t)$ for all $\Bx_t \in \T$.
\end{itemize}
Note that in the second condition, we use the fact that the kernel function
approximates the similarity between two instances~\cite{balcan2008theory}.
If we can find the most similar instance to each instance in $\R$,
the constraint $\sum_{s \in I_\Ss}{y_s \alpha_s'} + \sum_{t \in I_\T}{y_t \alpha_t'} = 0$ will be satisfied after
the replacing $\R$ by $\T$.
Whereas, if we cannot find any instance in $\T$ that has the same label as $\Bx_p$,
we randomly pick an instance from $\T$ to replace $\Bx_p$.
When the above situation happens,
the constraint $\sum_{s \in I_\Ss}{y_s \alpha_s'} + \sum_{t \in I_\T}{y_t \alpha_t'} = 0$ is violated.
Hence, we need to adjust $\Ba_\T'$ to make the constraint hold.
We use the same approach as MIR to adjusting $\Ba_\T'$.
The pseudo code for SIR is given in Algorithm~\ref{alg:sir} in Supplementary Material.

\eat{
To satisfy the constraint $\sum_{s \in I_\Ss}{y_s \alpha_s'} + \sum_{t \in I_\T}{y_t \alpha_t'} = 0$, we perform the following steps.
\begin{itemize}
	\item If $\sum_{t \in I_\T}{y_t \alpha_t'} > -\sum_{s \in I_\Ss}{y_s\alpha_s'}$,
		  we uniformly reduce all the $y_t \alpha_t'$ until $\sum_{t \in I_\T}{y_t \alpha_t'} = -\sum_{s \in I_\Ss}{y_s\alpha_s'}$,
		  subjected to the constraint $0 \le \alpha_t' \le C$.
	\item If $\sum_{t \in I_\T}{y_t \alpha_t'} < \sum_{s \in I_\Ss}{y_s\alpha_s'}$,
		  we uniformly increase all the $y_t \alpha_t'$ until $\sum_{t \in I_\T}{y_t \alpha_t'} = -\sum_{s \in I_\Ss}{y_s\alpha_s'}$,
		  subjected to the constraint $0 \le \alpha_t' \le C$.
\end{itemize}
After adjusting $\Ba_\T'$, Constraint~\eqref{eq:pro-const} holds.
We use $\Ba'$ (where $\Ba' = \Ba_\T' \cup \Ba_\Ss'$) as initialised alpha values to train the next SVM using SMO.
}

\begin{table*}
\vspace{-20pt}
\centering
\caption{Efficiency comparison ($k = 10$)}
\begin{adjustbox}{max width=1.0\textwidth}
\begin{tabular}{|*{14}{c|}} \hline
\multirow{3}{*}{Dataset}	& \multicolumn{7}{c|}{elapsed time (sec)}	
						& \multicolumn{4}{c|}{number of iterations}			
						& \multicolumn{2}{c|}{accuracy (\%)} \\\cline{2-14}
			& \multirow{2}{*}{libsvm} 	& \multicolumn{2}{c|}{ATO} & \multicolumn{2}{c|}{MIR}	& \multicolumn{2}{c|}{SIR}	
			& \multirow{2}{*}{libsvm}	& \multirow{2}{*}{ATO}	& \multirow{2}{*}{MIR}	& \multirow{2}{*}{SIR}
			&\multirow{2}{*}{libsvm}	&\multirow{2}{*}{SIR}\\\cline{3-8}
			& 			& init.	& the rest	& init. 	& the rest	& init. & the rest	& 			&	&		&			& 		& \\\hline
Adult 		& 6,783		& 3,824	& 5,738	& 2,034	& 3,717	& 57 	& 3,705 	& 397,565	& 361,914		& 318,169	& 317,110		& 82.36	& 82.36\\
Heart	 	& 0.36		& 0.016	& 0.19	& 0.058	& 0.083	& 0.003	& 0.24	& 6,988		& 4,882	& 1,443 	& 3,968		& 55.56	& 55.56 \\
Madelon	 	& 54.5		& 2.0	& 24.6	& 1.7	& 12.8	& 1.2	& 13.5	& 9,000		& 5,408	& 1,800		& 1,800		& 50	.0	& 50.0 \\
MNIST		& 172,816	& 35,410	\eat{est}& 69,435	\eat{est}& 30,897	& 38,696	& 1,416	& 36,406	& 1,291,068	& 575,250 \eat{est}& 280,820	& 258,500	& 50.85	& 50.85	\\
Webdata		& 24,689		& 11,166	& 9,394	& 6,172	& 7,574	& 133 	& 11,901& 783,208	& 245,385	& 230,357	& 356,528	& 97.70	& 97.70\\
\hline
\end{tabular}
\end{adjustbox}
\label{tbl:overall-eff}
\vspace{-15pt}
\end{table*}

\begin{table}[b]
\vspace{-25pt}
\centering
\caption{Datasets and kernel parameters}
\begin{small}
\begin{tabular}{|*{5}{c|}} \hline
Dataset	& Cardinality	& Dimension 		& $C$	& $\gamma$	\\\hline
Adult		& 32,561			& 123		& 100	& 0.5		\\
Heart	& 270			& 13			& 2182	& 0.2		\\
Madelon	& 2,000			& 500		& 1		& 0.7071		\\
MNIST	& 60,000			& 780		& 10		& 0.125		\\
Webdata & 49,749			& 300		& 64		& 7.8125		\\
\hline
\end{tabular}
\end{small}
\label{tbl:dataset}
\end{table}

\section{Experimental studies}
\label{paper:es}

We empirically evaluate our proposed algorithms using five datasets from the LibSVM website~\cite{chang2011libsvm}.
All our proposed algorithms were implemented in C++.
\eat{We used Eigen~\cite{eigenweb} to perform matrix operations (e.g. find the inverse of a matrix)
in Equations~\eqref{eq:delta-a-m} and~\eqref{eq:delta-alpha-appr}
for the ATO algorithm and MIR algorithm.
In our experiment, ATO breaks from the loop (i.e. lines 3-11 of Algorithm~\ref{alg:ato}),
if ATO executes the loop body for 100 times but $\R$ is still not empty.
}The experiments were conducted on a desktop computer running Linux with a 6-core E5-2620 CPU and 128GB main memory.
Following the common settings, we used the Gaussian kernel function and by default $k$ is set to $10$.
The hyper-parameters for each dataset are 
identical to the existing studies~\cite{catanzaro2008fast,smirnov2004unanimous,wufeature}.
Table~\ref{tbl:dataset} gives more details about the datasets.
We study the $k$-fold cross-validation under the setting of binary classification.
\eat{
Note that the MNIST problem is to predict the correct digit (ranging from 0 to 9) of a handwritten digit,
and it is a multi-class classification problem.
We converted it into a binary classification problem,
i.e. predicting whether a handwritten digit is an even or odd number~\cite{catanzaro2008fast}.
}

Next, we first show the overall efficiency of our proposed algorithms in comparison with LibSVM.
Then, we study the effect of varying $k$ from $3$ to $100$ in the $k$-fold cross-validation.
\eat{
and investigate the effect of varying the number of instances in a dataset.
Last, we present experimental results on leave-one-out cross-validation.
}

\subsection{Overall efficiency on different datasets}
We measured the total elapsed time of each algorithm to test their efficiency.
The total elapsed time consists of the alpha initialisation time and the time for the rest of the $10$-fold cross-validation.
The result is shown in Table~\ref{tbl:overall-eff}.
To make the table to fit in the page, we do not provide the total elapsed time of ATO, MIR and SIR for each dataset.
But the total elapsed time can be easily computed
by adding the time for alpha initialisation and the time for the rest.
\eat{
For example, the total elapsed time of ATO on Adult is adding the value (i.e. 3,824) in the third column
and the value (i.e. 5,738) in the fourth column of Table~\ref{tbl:overall-eff}.
}Note that the time for ``the rest" (e.g. the fourth column of Table~\ref{tbl:overall-eff})
includes the time for partitioning dataset into $10$ subsets, training (the most significant part) and classification.

As we can see from the table, the total elapsed time of MIR and SIR is much smaller than LibSVM.
In the Madelon dataset, MIR and SIR are about $2$ times and $4$ times faster than LibSVM, respectively.
In comparison, ATO does not show obvious advantages over MIR and SIR,
and is even slower than LibSVM on the Adult dataset due to spending too much time on alpha value initialisation.
Another observation from the table is SIR spent the smallest amount of time on the alpha initialisation
among our three algorithms, while SIR has the similar ``effectiveness'' as MIR on reusing the alpha values.
The effectiveness on reusing the alpha values is reflected by 
the total number of training iterations during the $10$-fold cross-validation.
More specifically, according to the ninth to twelfth columns of Table~\ref{tbl:overall-eff},
LibSVM often requires more training iterations than MIR and SIR;
SIR and MIR have similar number of iterations, and in some datasets (e.g. Adult and MNIST) SIR needs fewer iterations,
although SIR saves much time in the initialisation.
More importantly, the improvement on the efficiency does not sacrifice the accuracy. 
According to the last two columns of Table~\ref{tbl:overall-eff},
we can see that SIR produces the same accuracy as LibSVM.
Due to the space limitation, we omit providing the accuracy of ATO and MIR
which also produce the same accuracy as LibSVM.

\subsection{Effect of varying $k$}
We varied $k$ from $3$ to $100$ to study the effect of the value of $k$.
\eat{Please note that when $k = 2$, no training instances are shared between the two rounds of the $k$-fold cross-validation,
which violates the assumption of our alpha seeding techniques.
Hence, the minimum value for $k$ is $3$ for this set of experiments.
}Moreover, because conducting this set of experiments is very time consuming especially when $k = 100$,
we only compare SIR (the best among the our three algorithms according to results in Table~\ref{tbl:overall-eff})
with LibSVM.

Table~\ref{tbl:varying-k} shows the results.
Note that as LibSVM was very slow when $k=100$ on the MNIST dataset,
we only ran the first 30 rounds to estimate the total time.
As we can see from the table, SIR consistently outperforms LibSVM.
When $k=100$, SIR is about $32$ times faster than LibSVM in the Madelon dataset.
The experimental result for the leave-one-out (i.e. $k$ equals to the dataset size) cross-validation is similar to $k=100$,
and is available in Figure~\ref{fig:loocv} in Supplementary Material.

\begin{table}
\vspace{-8pt}
\caption{Effect of $k$ on total elapsed time (sec)}
\centering
\begin{adjustbox}{max width=0.48\textwidth}
\begin{tabular}{|*{7}{c|}} \hline
\multirow{2}{*}{Dataset}	& \multicolumn{2}{c|}{$k=3$}	
						& \multicolumn{2}{c|}{$k=10$}			
						& \multicolumn{2}{c|}{$k=100$} \\\cline{2-7}
			& libsvm		& SIR	& libsvm		& SIR	& libsvm		& SIR	 \\\hline
Adult 		& 733		&  683 	&  6,783		& 3,762	& 41,288		& 33,877	\\
Heart	 	& 0.09		& 0.08	& 0.36		& 0.25	& 3.39		& 1.17	 \\
Madelon	 	& 8.8		& 7.8	& 54.5		& 14.7	& 620		& 19.5	\\
MNIST		& 29,692		& 22,296	& 172,816	& 37,822	& 2,508,684	& 61,016	\\
Webdata		& 3,941 		& 2,342 	& 24,689		& 12,034	& 190,817	& 31,918	\\
\hline
\end{tabular}
\label{tbl:varying-k}
\end{adjustbox}
\vspace{-15pt}
\end{table}

\section{Related work}
\label{paper:rw}

We categorise the related studies into
two groups: on alpha seeding, and on online SVM training.

\subsection{Related work on alpha seeding}
DeCoste and Wagstaff~\cite{decoste2000alpha} first introduced the reuse of alpha values
in the SVM leave-one-out cross-validation.
Their method (i.e. AVG discussed in Supplementary Material) has two main steps: 
(i) train an SVM with the whole dataset; 
(ii) remove an instance from the SVM and distribute the associated alpha value uniformly among all the support vectors.
Lee et al.~\cite{lee2004efficient} proposed a technique (i.e. TOP discussed in Supplementary Material)
to improve the above method.
Instead of uniformly distributing alpha value among all the support vectors,
the method distributes the alpha value to the instance with the largest kernel value.

Existing studies called ``Warm Start"~\cite{kao2004decomposition,chu2015warm} apply alpha seeding in selecting the parameter $C$ for linear SVMs.
Concretely, $\Ba$ obtained from training the $h^{\text{th}}$ linear SVM with $C$ is used
for training the $h^{\text{th}}$ linear SVM with ($C+\Delta$) in the \textbf{two} $k$-fold cross-validation processes
by simply setting $\Ba' = r\Ba$ where $r$ is a ratio computed from $C$ and $\Delta$.
In those studies, no alpha seeding technique is used when training the $k$ SVMs with parameter $C$.
Our work aims to reuse the $h^{\text{th}}$ SVM for training the $(h+1)^{\text{th}}$ SVM
for the $k$-fold cross-validation with parameter $C$.

\subsection{Related work on online SVM training}
\eat{In the online SVM training, a subset of instances is outdated and removed, and meanwhile,
a subset of new instances arrives and is added. The online SVM training problem is similar to the $k$-fold cross-validation problem.
}
Gauwenberghs and Poggio~\cite{cauwenberghs2001incremental} introduced an algorithm for training SVM online
where the algorithm handles adding or removing one training instance.
Karasuyama and Takeuchi~\cite{karasuyama2009multiple} extended the above algorithm to the cases
where multiple instances need to be added or removed.
Their key idea is to gradually reduce the alpha values of the outdated instances to 0,
and meanwhile, to gradually increase the alpha values of the new instances.
Due to the efficiency concern, the algorithm produces \textit{approximate} SVMs.
Our work aims to train SVMs which meet the optimality condition.

\eat{
\subsection{Related work on accelerating $k$-fold cross-validation}

Some recent methods have been developed to improve the efficiency of SVM $k$-fold cross-validation using Graphic Processing Units (GPUs).
Athanasopouloset al.~\cite{athanasopoulos2011gpu} used GPUs
to precompute the kernel matrix, which is stored in main memory, to improve the efficiency of the SVM $k$-fold cross-validation.
Wen et al.~\cite{wen2014mascot} proposed to a more scalable GPU-based SVM $k$-fold cross-validation
by precomputing the kernel matrix and storing the kernel matrix to main memory extended by SSDs.
Our work aims to reuse the previously trained SVM for training the next SVM,
and our techniques can be integrated into the GPU-based algorithms.
The discussion on applying our techniques to the GPU-based algorithms is out of the scope of this paper
and hence is not provided here.
}

\section{conclusion}
\label{paper:conc}
To improve the efficiency of the $k$-fold cross-validation,
we have proposed three algorithms that reuse the previously trained SVM to initialise the next SVM,
such that the training process for the next SVM reaches the optimal condition faster.
We have conducted extensive experiments to validate the effectiveness and efficiency of our proposed algorithms.
Our experimental results have shown that the best algorithm among the three is SIR.
When $k=10$, SIR is several times faster than the $k$-fold cross-validation in LibSVM which does not make use of the previously trained SVM;
when $k=100$, SIR dramatically outperforms LibSVM (32 times faster than LibSVM in the Madelon dataset).
Moreover, our algorithms produce same results (hence same accuracy) as the $k$-fold cross-validation in LibSVM does.
\eat{
Our SIR algorithm can efficiently identify the support vectors and
accurately estimate their alpha values of the next SVM by using the previous SVM.
Hence, we recommend SIR for reusing the previous SVM.
}

\subsubsection{Acknowledgments}
This work is supported by Australian Research
Council (ARC) Discovery Project DP130104587 and Australian Research
Council (ARC) Future Fellowships Project FT120100832.
Prof. Jian Chen is supported by the Fundamental Research Funds for the Central Universities (Grant No. 2015ZZ029)
and the Opening Project of Guangdong Province Key Laboratory of Big Data Analysis and Processing.

\bibliographystyle{aaai}
\bibliography{wen-li}

\newpage
\section*{Supplementary Material}

\subsection{Pseudo-code of our three algorithm}
Here, we present the pseudo-code of our three algorithms proposed in the paper.
\subsubsection{The ATO algorithm}
The full algorithm of ATS is summarised in Algorithm~\ref{alg:ato}.
As we can see from Algorithm~\ref{alg:ato}, ATO terminates when $\R$ is empty
and it might spend a substantial time in the loop especially when the step size $\eta$ is small.
\begin{algorithm}
\begin{small}
\KwIn{Sets $\X$ and $\R$ of instances,\\
\quad \quad \quad  $\Ba$ associated with instances in $\X$,\\
\quad \quad \quad  and a set $\T$ of new instances.}
\KwOut{Optimal alpha values for $\X \setminus \R$ and $\T$.}

$\Ba'_\T \leftarrow \boldsymbol{0}$\tcc*[f]{Initialise $\Ba'_\T$}\\
\tcc*[f]{Initialise index sets $I_m$, $I_u$ and $I_l$}\\
Init($I_m$, $I_u$, $I_l$, $\Ba$)\\
\Repeat{$\R = \phi$}{
	$\eta \leftarrow$ GetStepSize() \tcc*[f]{Eqs~\eqref{eq:delta-a-rt}, \eqref{eq:delta-a-m} and~\eqref{eq:delta-f}}\\
	\tcc*[f]{use Eqs~\eqref{eq:delta-a-rt} and~\eqref{eq:delta-a-m} to update $\boldsymbol{\alpha}$} \text{\hspace{6pt}}\\
	$\Ba'$, $\Ba'_\T \leftarrow$
	UpdateAlpha($\eta$, $\Ba$, $\Ba'_\T$)
	
	$\Bf \leftarrow$ UpdateF($\eta$, $\Bf$)\tcc*[f]{use Eqs~\eqref{eq:f-i} and~\eqref{eq:delta-f}}

	\ForEach{$r \in I_\R$}
	{
		\If(\tcc*[f]{safe to remove $\Bx_r$}){$\alpha_r' = 0$}
		{
			$I_\R \leftarrow I_\R \setminus \{r\}$, $\Ba' \leftarrow \Ba' \setminus \{\alpha_r'\}$
		}
	}
	
	\tcc*[f]{update the sets $I_m$, $I_g$ and $I_s$}\\
	$I_m$, $I_u$, $I_l \leftarrow$
	Rearrange($I_m$, $I_u$, $I_l$, $\Bf$)	
}
$\Ba' \leftarrow \Ba'_\T \cup \Ba'$\\
$\X' \leftarrow \T \cup \X \setminus R$\\
TrainOptimalSVM($\Ba'$, $\X'$)\tcc*[f]{SMO to improve $\Ba'$}

\caption{Adjusting Alpha Towards Optimum (ATO)}
\label{alg:ato}
\end{small}
\end{algorithm}

\subsubsection{The MIR algorithm}
The full algorithm of MIR is summarised in Algorithm~\ref{alg:rt-replacement}.
\begin{algorithm}
\begin{small}
\KwIn{Sets $\X$ and $\R$ of instances,\\
\quad \quad \quad  $\Ba$ associated with instances in $\X$,\\
\quad \quad \quad  and a set $\T$ of new instances.}
\KwOut{Optimal alpha values for $\X \setminus \R$ and $\T$.}

$\Ba'_\T \leftarrow \boldsymbol{0}$\tcc*[f]{Initialise $\Ba'_\T$ for $\T$}\\
\tcc*[f]{Initialise index sets $I_m$, $I_u$ and $I_l$}\\
Init($I_m$, $I_u$, $I_l$, $\Ba$)\\

	$\Delta f_i \leftarrow$ ComputeDeltaF() \tcc*[f]{Equation~\eqref{eq:delta-f-const}}\\
	$\Ba'_\T \leftarrow$ ComputeAlpha($\Ba_\R$, $\By_\R$) \tcc*[f]{Equation~\eqref{eq:delta-alpha-appr}}\\
	\tcc*[f]{Adjust $\Ba'_\T$ to meet constraints of problem~\eqref{eq:svm_dual}}\\
	$\Ba'_\T \leftarrow$ AdjustAlpha($\Ba'_\T$, $\By_\T$, $\Ba_\R$, $\By_\R$) 

$\Ba' \leftarrow \Ba'_\T \cup \Ba \setminus \Ba_\R$\\
$\X' \leftarrow \T \cup \X \setminus R$\\
TrainOptimalSVM($\Ba'$, $\X'$)\tcc*[f]{SMO to improve $\Ba'$}

\caption{Multiple Instance Replacement (MIR)}
\label{alg:rt-replacement}
\end{small}
\end{algorithm}

\subsubsection{The SIR algorithm}
The full algorithm of SIR is summarised in Algorithm~\ref{alg:sir}.

\begin{algorithm}
\begin{small}
\KwIn{Sets $\X$ and $\R$ of instances,\\
\quad \quad \quad  $\Ba$ associated with instances in $\X$,\\
\quad \quad \quad  and a set $\T$ of new instances.}
\KwOut{Optimal alpha values for $\X \setminus \R$ and $\T$.}

$\Ba'_\T \leftarrow \boldsymbol{0}$\tcc*[f]{Initialise $\Ba'_\T$ for $\T$}\\
	
\ForEach{$r \in I_\R$}
{
	$maxValue \leftarrow 0$, $t' \leftarrow -1$\\
	\ForEach{$t \in I_\T$}
	{
		\If{$y_r = y_t \land K(\Bx_r, \Bx_t) > maxValue$}
		{
			$maxValue \leftarrow K(\Bx_r, \Bx_t)$, $t' \leftarrow t$
		}
	}
	\If(\tcc*[f]{replace $\Bx_r$ by $\Bx_{t'}$}){$t' \neq -1$}
	{
		$I_\T \leftarrow I_\T \setminus \{t'\}$, $\Ba \leftarrow \Ba \setminus \{\alpha_r\}$, $\alpha_{t'} \leftarrow \alpha_r$
	}
}

	\tcc*[f]{Adjust $\Ba'_\T$ to meet constraints of Problem~\eqref{eq:svm_dual}}\\
	$\Ba'_\T \leftarrow$ AdjustAlpha($\Ba'_\T$, $\By_\T$, $\Ba$, $\By_\Ss$) 

$\Ba' \leftarrow \Ba'_\T \cup \Ba$\\
$\X' \leftarrow \T \cup \X \setminus R$\\
TrainOptimalSVM($\Ba'$, $\X'$)\tcc*[f]{SMO to improve $\Ba'$}

\caption{Single Instance Replacement (SIR)}
\label{alg:sir}
\end{small}
\end{algorithm}

\subsection{Existing approaches for leave-one-out cross-validation}
\label{paper:alg:loocv}

As our three algorithms (i.e. ATO, MIR and SIR) are proposed to improve the efficiency of $k$-fold cross-validation,
naturally the three algorithms can accelerate leave-one-out cross-validation.
Note that leave-one-out cross-validation is a special case of $k$-fold cross-validation,
when $k$ equals to the number of instances in the dataset.
Here, we present two existing alpha seeding techniques~\cite{decoste2000alpha,lee2004efficient}
that have been specifically proposed to improve the efficiency
of leave-one-out cross-validation.

Given a dataset $\X$ of $n$ instances,
both of the algorithms train the SVM using all the $n$ instances.
Recall that the trained SVM meets constraints of Problem~\eqref{eq:svm_dual},
and we have the constraint $\sum_{\Bx_i \in \X}{y_i \alpha_i} = 0$ held.
Then, in each round of the leave-one-out cross-validation, an instance $\Bx_t$ is removed from the trained SVM.
To make the constraint $\sum_{\Bx_i \in \X \setminus \{\Bx_t\}}{y_i \alpha_i'} = 0$ hold,
the alpha values of the instances in $\X \setminus \{\Bx_t\}$ may need to be adjusted.
The two existing techniques apply different strategies to adjust the alpha values of the instances in $\X \setminus \{\Bx_t\}$.

\subsubsection{Uniformly distributing $\alpha_t y_t$ to other instances}
First, the strategy proposed in~\cite{decoste2000alpha} counts the number,
denoted by $d$, of instances with alpha values satisfying $0 < \alpha_i < C$ where $\Bx_i \in \X\setminus \{\Bx_t\}$.
Then, the average amount of value that the $d$ instances need to be adjusted is $\frac{y_t \alpha_t}{d}$.
For each instance $\Bx_j$ in the $d$ instances, adjusting their alpha values is handled in
the following two scenarios.
\begin{itemize}
	\item If $y_t = y_j$, $\alpha_j'$ equals to $(\alpha_j + \frac{\alpha_t}{d})$.
	\item If $y_t = -y_j$, $\alpha_j'$ equals to $(\alpha_j - \frac{\alpha_t}{d})$.
\end{itemize}
Note that the updated alpha value $\alpha_j'$ subjects to the constraint $0 \le \alpha_j' \le C$.
Hence, the alpha values of some instances may not allow to be increased/decreased by $\frac{\alpha_t}{d}$.
Those alpha values are adjusted to the maximum allowed limit (i.e. increased to $C$ or decreased to $0$);
similar to the above process, the extra amount of value
(of $\frac{\alpha_t}{d}$ which cannot be added to or removed from $\alpha_j'$)
is uniformly distributed to those alpha values that satisfy $0 < \alpha_i < C$.

We call this technique \textbf{AVG}, because each alpha value of the $d$ instances is increased/decreased
by the average amount (except those near $0$ or $C$) of value from $y_t \alpha_t$.
Our ATO algorithm has the similar idea as AVG,
where the alpha values of many instances are adjusted by the same (or similar) amount.

\subsubsection{Distributing the $\alpha_t y_t$ to similar instances}

AVG requires changing the alpha values of many instances, which may not be efficient.
Lee et al.~\cite{lee2004efficient} proposed a technique to adjust the alpha values of only a few most similar instances to $\Bx_t$.
The technique first finds the instance $\Bx_j$ among $\X \setminus \{\Bx_t\}$ with the largest kernel value,
i.e. $K(\Bx_j, \Bx_t)$ is the largest.
Then, $\alpha_j' \leftarrow (\alpha_j + \alpha_t)$ if $y_t = y_j$ or $\alpha_j' \leftarrow (\alpha_j - \alpha_t)$ if $y_t = -y_j$.
Recall that the updated alpha value $\alpha_j'$ needs to satisfy the constraint $0 \le \alpha_j' \le C$.
Hence, the alpha value of the most similar instance $\Bx_j$ may not allow to be increased/decreased by $\alpha_t$.
Then, $\alpha_j'$ is increased to $C$ or decreased to $0$ depending on $y_j$.
The extra amount of value is distributed to the alpha value of the second most similar instance,
the third most similar instance, and so on
until the constraint $\sum_{\Bx_i \in \X \setminus \{\Bx_t\}}{\alpha_i' y_i} = 0$ holds.

We call this technique \textbf{TOP}, since it only adjusts the alpha values of a few most similar (i.e. a top few) instances to $\Bx_t$.
Our MIR algorithm and SIR algorithm have the similar idea to TOP,
where only the alpha values of a proportion of the instances are adjusted.

After the adjusting by either of the two techniques,
the constraint $\sum_{\Bx_i \in \X \setminus \{\Bx_t\}}{\alpha_i' y_i} = 0$ holds,
and $\Ba'$ is used as the initial alpha values for training the next SVM.
In the next section (more specifically, in Section~\ref{paper:loo-exp}),
we empirically evaluate the five techniques for accelerating leave-one-out cross-validation.

\subsection{Efficiency comparison on leave-one-out cross-validation}
\label{paper:loo-exp}

Here, we study the efficiency of our proposed algorithms,
in comparison with LibSVM and the existing alpha seeding techniques, i.e. AVG and TOP (cf. Section~\ref{paper:alg:loocv}),
for leave-one-out cross-validation.
Similar to the other algorithms, we implemented AVG and TOP in C++.
Since leave-one-out cross-validation is very expensive for the large datasets,
we estimated the total time for leave-one-out cross-validation on the three large datasets (namely Adult, MNIST and Webdata) for each algorithm.
For MNIST and Webdata, we ran the first 30 rounds of the leave-one-out cross-validation to estimate the total time for each algorithm;
for Adult, we ran the first 100 rounds of the leave-one-out cross-validation to estimate the total time for each algorithm.
As Heart and Madelon are relatively small, we ran the whole leave-one-out cross-validation, and measured their total elapsed time.
The experimental results are shown in Figure~\ref{fig:loocv}.
As we can see from the table, all the five algorithms
are faster than LibSVM ranging from a few times to a few hundred times (e.g. SIR is 167 times faster than LibSVM on Webdata).
Another observation from the table is AVG and TOP have similar efficiency.
It is worth pointing out that our SIR algorithm almost always outperforms all the other algorithms,
except Heart and Madelon where MIR is slightly better.

\eat{
From the above experimental study, we recommend SIR to accelerate $k$-fold cross-validation,
because SIR has the following three advantages.
(i) SIR is generally more efficient than other algorithms on various datasets (cf. Table~\ref{tbl:overall-eff});
(ii) SIR is robust while varying $k$ (cf. Table~\ref{tbl:varying-k});
(iii) SIR has better scalability over the dataset size (cf. Figure~\ref{fig:varying-size}).
}

\begin{figure}
\center
\includegraphics[width=3in, height=3in]
{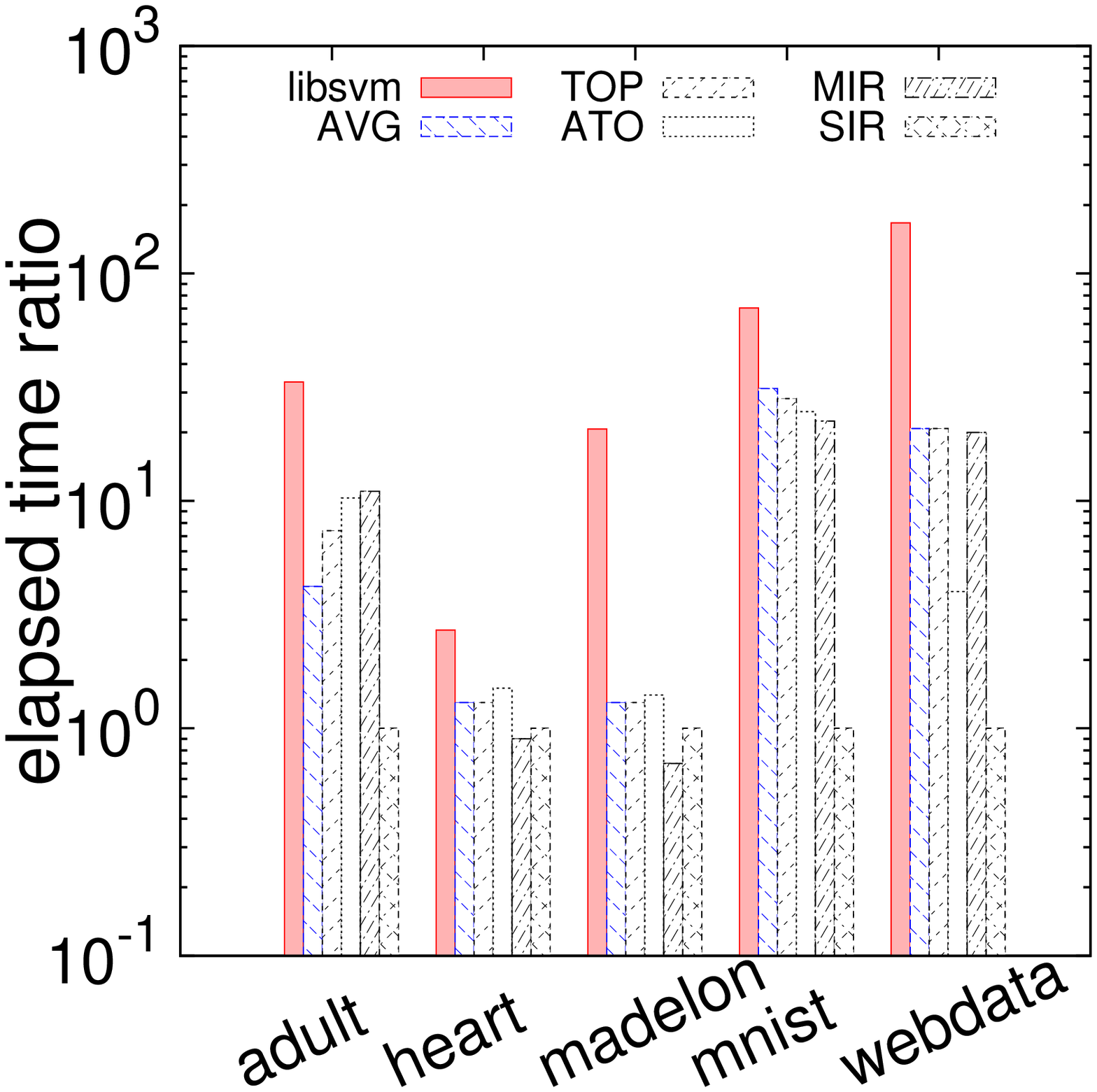}
\vspace{-55pt}
\caption{Elapsed time compared with the total elapsed time of SIR in leave-one-out cross-validation}
\label{fig:loocv}
\end{figure}

\end{document}